\documentclass[letterpaper, 10pt, conference]{ieeeconf}
\IEEEoverridecommandlockouts 
\usepackage{amsfonts,amssymb,amsmath,mathtools,multicol}
\usepackage{physics}
\usepackage{cancel}

\usepackage{float}
\usepackage{graphicx}
\graphicspath{{figures/}}
\usepackage{subfig}

\usepackage{booktabs}   
\usepackage{multirow}
\usepackage{diagbox}
\usepackage{makecell}
\usepackage{subcaption}

\usepackage{algorithm}
\usepackage[noend]{algpseudocode}

\usepackage[bibstyle=ieee,citestyle=numeric-comp,sortcites,mincitenames=1,maxcitenames=2,natbib=true,backend=biber]{biblatex}
\AtBeginBibliography{\small}
\usepackage[hidelinks]{hyperref}

\usepackage[dvipsnames,table,xcdraw]{xcolor}
\usepackage{soul}
\usepackage[normalem]{ulem}

\sethlcolor{yellow}

\newcommand{\R}{\mathbb{R}} 
\newcommand{\E}{\mathbb{E}} 
\newcommand{\V}[1]{\boldsymbol{\mathbf{#1}}} 
\newcommand{\T}{\top}
\newcommand{\VT}[1]{\V{#1}^\T}

\newcommand{\Vbar}[1]{\bar{\V{#1}}}
\newcommand{\Vhat}[1]{\hat{\V{#1}}}

\newcommand{\bmat}[1]{\bmqty{#1}}

\usepackage[acronym]{glossaries}
\setacronymstyle{long-short}
\newcommand{\g}[1]{\gls[hyper=false]{#1}} 
\newacronym{LfD}{LfD}{Learning from Demonstration}
\newacronym{GMM}{GMM}{Gaussian Mixture Model}
\newacronym{GMR}{GMR}{Gaussian Mixture Regression}
\newacronym{TP-GMM}{TP-GMM}{task-parameterized Gaussian Mixture Model}
\newacronym{DMP}{DMP}{Dynamic Movement Primitives}
\newacronym{ProMP}{ProMP}{Probabilistic Movement Primitives}
\newacronym{RRT}{RRT}{Rapidly-exploring Random Tree}
\newacronym{EM}{EM}{Expectation-Maximization}
\newacronym{SPD}{SPD}{symmetric positive definite}

\title{\textbf{Generalizing Robot Trajectories from Single-Context Human Demonstrations: A Probabilistic Approach}}

\author{Qian Ying Lee$^1$, Suhas Raghavendra Kulkarni$^1$, Kenzhi Iskandar Wong$^1$, Lin Yang$^1$,\\Bernardo Noronha$^1$, Yongjun Wee$^2$, Domenico Campolo$^{1,\ast}$
\thanks{This research was conducted under project WP3 within the Delta-NTU Corporate Lab with funding support from A*STAR under its IAF-ICP programme (Grant no. I2201E0013) and Delta Electronics Inc.}
\thanks{$^{1}$Robotics Research Centre, School of Mechanical
and Aerospace Engineering, Nanyang Technological University, Singapore.}%
\thanks{$^{2}$Delta-NTU Corporate Lab, Singapore.}%
\thanks{$^{\ast}$Corresponding author: {\tt\small d.campolo@ntu.edu.sg}}
}

\begin{document}
\maketitle
\thispagestyle{empty}
\pagestyle{empty}

\begin{abstract}
Generalizing robot trajectories from human demonstrations to new contexts remains a key challenge in \g{LfD}, particularly when only single-context demonstrations are available. We present a novel \g{GMM}-based approach that enables systematic generalization from single-context demonstrations to a wide range of unseen start and goal configurations. Our method performs component-level reparameterization of the \g{GMM}, adapting both mean vectors and covariance matrices, followed by \g{GMR} to generate smooth trajectories. We evaluate the approach on a dual-arm pick-and-place task with varying box placements, comparing against several baselines. Results show that our method significantly outperforms baselines in trajectory success and fidelity, maintaining accuracy even under combined translational and rotational variations of task configurations. These results demonstrate that our method generalizes effectively while ensuring boundary convergence and preserving the intrinsic structure of demonstrated motions.
\end{abstract}
\glsresetall

\section{Introduction}
\g{LfD} provides a powerful framework for enabling robots to acquire complex manipulation skills by observing and imitating human behavior, aiming to transfer human expertise and improve adaptability in robotic task execution \cite{argall2009survey, ravichandar2020recent, si2021review}. It is particularly advantageous for tasks that are hard to specify through explicit programming, especially when task parameters or environmental conditions vary. As a result, it has been widely adopted in a wide range of application domains such as manufacturing, warehouse automation, and service robotics.

Effective \g{LfD} requires parameterizing demonstration data to encode observed behavior into a compact and generalizable representation of the underlying policy \cite{calinon2009robot}. This involves two complementary components: (\textit{i}) task-specific parameters, capturing contextual information relevant to task execution, such as object configurations and interaction constraints; and (\textit{ii}) motion primitive parameters, describing the spatiotemporal aspects of the behavior, including trajectory geometry, timing, and kinematic or dynamic properties. Several frameworks encode these parameters to enable reproduction of the demonstrated behavior. \g{GMM} \cite{calinon2007learning} models trajectory data probabilistically, while \g{GMR} \cite{cohn1996active} enables smooth interpolation. \g{DMP} \cite{ijspeert2013dynamical} models trajectories as nonlinear differential equations with stable attractors, enabling temporal and spatial modulation while preserving motion shape. Similarly, \g{ProMP} \cite{paraschos2013probabilistic} represent trajectories as distributions over basis functions, supporting uncertainty-aware reproduction and conditioning on partial observations.

While these frameworks effectively learn and reproduce motions from demonstrations, generalizing to new task contexts remains challenging \cite{osa2018algorithmic}. Task-parameterized models, such as \g{TP-GMM} \cite{calinon2016tutorial}, address this by incorporating contextual information into learning. \g{TP-GMM} extends standard \g{GMM} by representing demonstrations in multiple task-relevant reference frames, allowing the learned behavior to adapt to variations in task parameters like object poses. The global trajectory distribution is obtained by combining Gaussians from each local frame, projected into the global frame based on the current task configuration. Empirical results \cite{zhu2022learning} suggest that diverse initial demonstrations improve \g{TP-GMM} generalization. Consequently, the framework relies on demonstrations spanning a wide range of conditions, referred in this paper as \textit{multi-context demonstrations}. Excessive variability in demonstrations can introduce ambiguities and degrade model fidelity \cite{franzese2020learning}, which recent studies mitigate via synthetic data augmentation \cite{zhu2022learning, prados2024learning}. Additionally, \g{TP-GMM} struggles to guarantee precise endpoint convergence during generalization. To address this, \citet{sena2019improving} proposed weighting frames with lower covariance, thereby allowing high-precision frames to dominate, reducing distortion in global \g{GMM} and improves extrapolative generalization and endpoint accuracy. Nonetheless, it still relies on multi-context demonstrations, and no existing method reliably ensures convergence at boundary points (start and goal) under interpolation and extrapolation scenarios.

While \g{TP-GMM} addresses some generalization challenges by using multiple reference frames, generalization in \g{LfD} remains a broader issue, often requiring multi-context demonstrations that may be impractical in real-world settings. In comparison, the standard \g{GMM} operates in a single reference frame, directly modeling the spatiotemporal distribution of demonstration data. While simpler, it is mainly designed for trajectory reproduction and does not inherently support generalization to new task contexts. \citet{kyrarini2019robot} attempted to generalize standard \g{GMM} to new goal positions by displacing component means while keeping covariance matrices fixed. As a result, for large or opposing displacements, \g{GMR}-generated trajectories may exhibit discontinuities, fail to retain the skill-specific structure inherent in the demonstrations, and may ultimately lead to task failure.

Current approaches often focus on generating feasible trajectories that achieve the task goals, but they typically overlook the preservation of the nuanced aspects of human behavior in the demonstrations. For instance, \g{DMP}s reproduce overall trajectory shape and are robust to perturbations but smooth out fine-grained motion details. \g{ProMP}s also capture the general trajectory shape while additionally modeling variability and coordination across multiple demonstrations. However, their representation is still limited by the training data distribution, making extrapolation to new contexts or preservation of structural coherence characteristic challenging. In both cases, the emphasis remains on trajectory reproduction. These limitations motivate the need for methods capable of extracting task-invariant features from demonstrations and adapting them precisely across diverse configurations, even under single-context demonstrations.

In this work, we propose a novel \g{GMM}-based algorithm that generalizes from single-context demonstrations, i.e., where start and goal poses are nearly identical, to a wide range of previously unseen task scenarios, encompassing both interpolation and extrapolation. Our approach ensures convergence to task boundary points while preserving the intrinsic characteristics of the demonstrated skills throughout execution. The main contributions of this work are:
\begin{enumerate}
    \item \textbf{Component-level reparameterization of \g{GMM} for generalization}, a novel method that adapts both the means and covariance matrices of a \g{GMM} learned from single-context demonstrations to new task configurations (start and end states).
    \item \textbf{Convergence to boundary points}, ensuring that generalized motions respect both the start and end configurations under novel task conditions.
    \item \textbf{Preservation of skill-specific motion structure}, ensuring that the essential patterns of the demonstrated motion are maintained during generalization.
\end{enumerate}

\section{Method}\label{sec: method}
In this section, we present a novel approach for systematic generalization of \g{GMM} representations learned from single-context demonstrations. The method has three main stages:
\begin{enumerate}
    \item \textbf{Skill encoding via \g{GMM}} (Sec. \ref{sec: gmm_training}): Demonstrations are modeled as a Gaussian mixture, capturing both the variability and underlying structure of the behavior.
    \item \textbf{Component-level reparameterization} (Sec. \ref{sec: gmm_generalization}): Adapts the learned model to new task contexts while retaining the key characteristics of the demonstration.
    \item \textbf{Trajectory generation} (Sec. \ref{sec: regression}): \g{GMR} is used to produce smooth trajectories.
\end{enumerate}

\subsection{Gaussian Mixture Model (GMM)}\label{sec: gmm}
To establish the conceptual foundation of our approach, we first review the core theoretical principles of \g{GMM} \cite{calinon2007learning}, which form the theoretical basis of our methodology.

A \g{GMM} with $G$ Gaussian components is characterized by the parameter set $\{\pi_g, \V\mu_g, \V\Sigma_g\}_{g=1}^G$, where each $\pi_g$ denotes a non-negative mixing coefficient satisfying $\pi_g > 0$ and $\sum_{g=1}^G \pi_g = 1$. The parameters $\V\mu_g$ and $\V\Sigma_g$ denote the mean (center) vector and the \g{SPD} covariance matrix of the $g$-th component, respectively.

To model the underlying distribution of human demonstrations $\V\xi$, the \g{GMM} approximates the joint probability density as a convex combination of multivariate Gaussian components \cite{verbeek2004mixture}:
\begin{equation}
    p(\V\xi) = \sum_{g=1}^G \pi_g \mathcal{N}(\V\xi \mid \V\mu_g,\V\Sigma_g),
\end{equation}
where $\V\xi \in \R^{D+1}$ with $D$ denotes the data dimensionality, and the additional dimension encodes the progression of the demonstration (e.g. time).\\
The multivariate Gaussian distribution is then given by:
\begin{equation}
    \mathcal{N}(\V\xi \mid \V\mu_g,\V\Sigma_g) = \frac{\exp(-\frac{1}{2} (\V\xi - \V\mu_g)^\T\V\Sigma_g^{-1}(\V\xi - \V\mu_g))}{(2\pi)^{(D+1)/2} |\V\Sigma_g|^{1/2}}.
\end{equation}

The optimal parameters of the \g{GMM} are estimated using the \g{EM} algorithm \cite{dempster1977maximum, mclachlan2008algorithm}, which performs maximum likelihood estimation iteratively. In the expectation (E) step, the posterior probabilities of the latent variables are computed given the current parameter estimates. In the maximization (M) step, the model parameters $\{\pi_g, \V\mu_g, \V\Sigma_g\}$ are updated to maximize the expected log-likelihood. This iterative procedure typically converges toward a local maximum of the likelihood function.

\subsection{Skill Encoding via \g{GMM}}\label{sec: gmm_training}
Human demonstrations of manipulation tasks typically comprise a sequence of distinct motor skills, such as grasping, transporting, and reorienting objects, collectively capturing the complexity of human behavior. Effectively modeling such behavior requires representing the variability inherent in demonstrations, which improves movement fidelity, enhances task precision, and enables adaptive performance. In this context, \g{GMM}s are a powerful tool for capturing the statistical structure of complex manipulation behaviors.

Let $\mathcal{J}$ denote a set of human demonstrations. Each demonstration $j \in \mathcal{J}$ is represented as a time series $t \mapsto \V x_j(t) \in \R^D$, where $\V x_j(t)$ is the pose state at time $t$, and $D$ is the data dimensionality. The complete trajectory of demonstration $j$ is denoted as $\V\xi_j = \bmat{t & \V x_j(t)^\T}^\T \in \R^{D+1}$, with $t \in [0, T]$ for all $j \in \mathcal{J}$, and $T$ denotes the duration of the demonstrations.

The set of demonstrations $\{\V\xi_j\}_{j \in \mathcal{J}}$ is first clustered into $G$ Gaussian components using K-means algorithm \cite{macqueen1967multivariate}, followed by parameter optimization via the \g{EM} algorithm. The resulting \g{GMM} parameters, including component priors $\pi_g \in \R$, mean vectors $\V\mu_g$, and covariance matrices $\V\Sigma_g$, are expressed as:
\begin{align}
    \V\mu_g &= \bmat{\bar{t}_g & \Vbar x_g^\T}^\T \in\R^{D+1}, \\
    \V\Sigma_g &= \bmat{\Sigma_{tt,g} & \V\Sigma_{tx,g} \\ \V\Sigma_{xt,g} & \V\Sigma_{xx,g}} \in\R^{(D+1)\times(D+1)}. \label{eq: cov_matrix}
\end{align}
where $\bar{(\cdot)}$ denotes the mean of the corresponding variable.

\begin{figure}[t]
    \centering
    \includegraphics[width=0.48\textwidth]{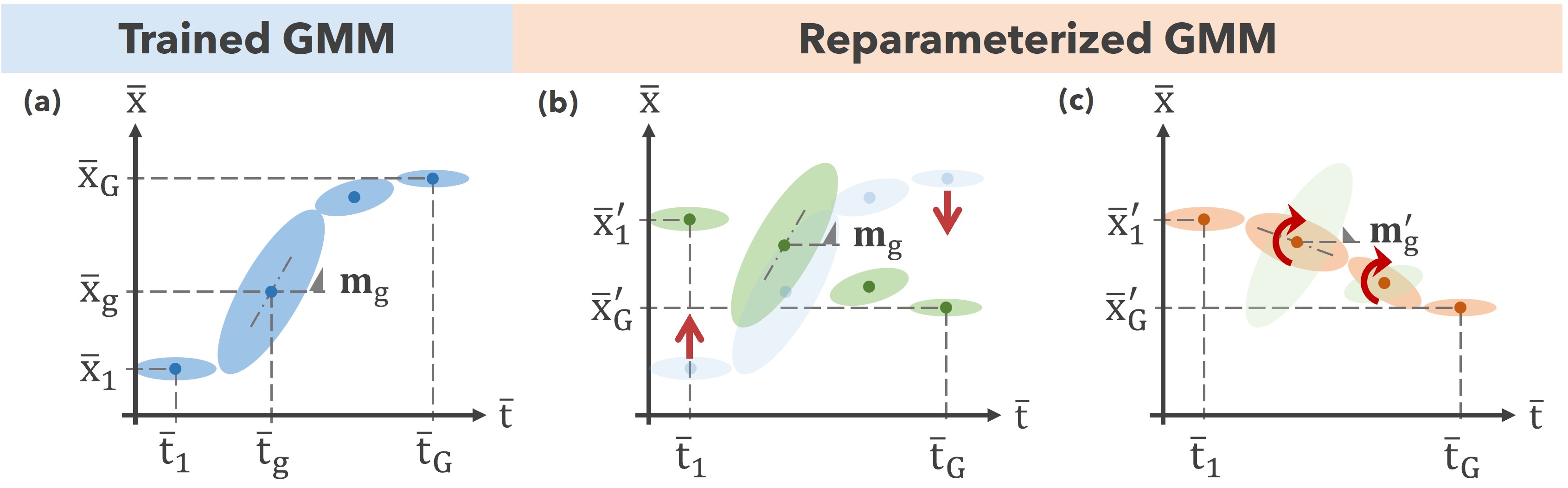}
    \caption{Gaussian components capturing spatiotemporal correlations (single dimension): (a) original \g{GMM} trained on demonstration data, (b) after mean vectors reparameterization, and (c) after covariance matrices reparameterization.}
    \label{fig: gmm_illustration}
\end{figure}

Fig. \ref{fig: gmm_illustration}a shows Gaussian components that compactly encode the spatiotemporal correlations in the demonstrations.

\subsection{Component-Level Reparameterization of \g{GMM}}\label{sec: gmm_generalization}
While \g{GMM}-based skill encoding enables precise reproduction of demonstrated behaviors, the model often struggles to generalize to unseen scenarios. In this context, the parameters $\{\pi_g,\V\mu_g,\V\Sigma_g\}$ provide sufficient statistics for reproduction, suggesting that targeted adjustments to the means $\V\mu_g$ and covariance matrices $\V\Sigma_g$ can enhance performance in novel contexts. Motivated by this, we propose a component-level adaptation algorithm to systematically adjust these parameters for new task configurations.

Consider a \g{GMM} with $G$ Gaussian components encoding demonstration data over time in a continuous spatial domain, illustrated in Fig. \ref{fig: gmm_illustration}a (shown along a single spatial dimension for clarity). Assuming fixed movement timing and mixing coefficients ($\bar t_g$, $\Sigma_{tt,g}$, and $\pi_g$), we outline a two-stage transformation procedure (Fig. \ref{fig: gmm_illustration}b and \ref{fig: gmm_illustration}c) adapting the Gaussian components. These transformations are conditioned solely on two external vectors, $\Vbar x'_1$ and  $\Vbar x'_G$, specifying the updated first and last Gaussian components, respectively. All other components are derived from the model. \vspace{0.4em}\\
\textbf{(i) Reparameterization of mean vectors}. Let $\Vbar x'_g$ denote the updated mean vector of the $g$-th component. Given the updated mean vectors of the first and last Gaussian components, $\Vbar x'_1$ and $\Vbar x'_G$, the mean vector $\Vbar x'_g$ for any intermediate component $g$ is computed as:
\begin{equation}
    \Vbar x'_g = \Vbar x'_1 + \Delta\Vbar X'_G \Delta\Vbar X_G^{-1} \Delta\Vbar x_g,\quad g = 1,\dots,G
\end{equation}
where 
\begin{align}
    \Delta\Vbar X'_G &= \operatorname{diag}(\Vbar x'_G - \Vbar x'_1)\\
    \Delta\Vbar X_G^{-1} &= \operatorname{diag}(\Vbar x_G - \Vbar x_1)^{-1}\\
    \Delta\Vbar x_g &= \Vbar x_g - \Vbar x_1
\end{align}
Here, $\operatorname{diag}(\cdot)$ denotes the diagonal matrix formed from the vector differences, applied element-wise. \vspace{0.4em}\\
\textbf{(ii) Reparameterization of covariance matrices}. The updated Gaussian components shown in Fig. \ref{fig: gmm_illustration}b require refinement to ensure a smooth trajectory during regression-based reconstruction. In this context, the covariance matrices characterize the geometry of the Gaussian components, including their orientation and scale in the multidimensional feature space. Therefore, reparameterizing the covariance matrices is essential for a smooth and coherent reconstructed trajectory.

Under the assumption of the fixed $\Sigma_{tt,g}$, the covariance matrix in \eqref{eq: cov_matrix} can be reformulated as:
\begin{equation}
    \V\Sigma_g = \Sigma_{tt,g} \bmat{1 & \VT m_g \\ \V m_g & \V C_g}
\end{equation}
where $\V m_g = \V\Sigma_{xt,g} \Sigma_{tt,g}^{-1}\in\R^D$ is the normalized direction of the $g$-th component. Geometrically, it corresponds to the local `slope', that is the deviation of the component from its center $\Vbar x_g$ per unit time step. The term $\V C_g = \V\Sigma_{xx,g}\Sigma_{tt,g}^{-1} \in\R^{D \times D}$ is the normalized spatial covariance matrix of the $g$-th component.\\
Thus, the updated covariance matrix $\V\Sigma'_g$ can be obtained by modifying the parameters $\V m_g$ and $\V C_g$. The newly normalized direction of the $g$-th component is given by:
\begin{equation}
    \V m'_g = \Delta\Vbar X' \Delta\Vbar X^{-1} \V m_g,\quad g=2,\dots,G
\end{equation}
where
\begin{align}
    \Delta\Vbar X' &= \operatorname{diag}(\Vbar x'_g - \Vbar x'_{g-1})\\
    \Delta\Vbar X^{-1} &= \operatorname{diag}(\Vbar x_g - \Vbar x_{g-1})^{-1}
\end{align}
To ensure that the updated covariance matrix $\V\Sigma'_g$ remains \g{SPD}, the Schur complement is applied:
\begin{equation}
\V C'_g - \V m'_g \V m'^\T_g \succ 0,
\end{equation}
ensuring that the remaining covariance after accounting for the mean contribution is positive definite. The covariance block can then be efficiently updated via a rank-two update:
\begin{equation}
\V C'_g = \V C_g + \V m'_g \V m'_g{}^\T - \V m_g \VT m_g,
\end{equation}
where the difference of two rank-one matrices replaces the old mean contribution with the new one. Together, this approach ensures that $\V\Sigma'_g$ remains \g{SPD} while incorporating the updated mean efficiently.

Fig. \ref{fig: gmm_illustration}c depicts the reparameterized \g{GMM} obtained through the two-stage transformations.

\subsection{Gaussian Mixture Regression (GMR)}\label{sec: regression}
To reconstruct the motion from the reparameterized \g{GMM} parameters, \g{GMR} \cite{calinon2016tutorial} is used to estimate the pose $\Vhat x$ at a given time $t$ by conditioning on the joint probability distribution. Specifically, the regressed pose at time $t$ is given by:
\begin{equation}\label{eq: gmr}
\Vhat x(t) = \E[\V x|t] = \sum_{g=1}^G h_g(t) [\Vbar x'_g + \V m'_g(t - \bar{t}_g)],
\end{equation}
where
\begin{equation}
h_g(t) = \frac{\pi_g \mathcal{N}(t \mid \bar{t}_g,\Sigma_{tt,g})}{\sum_{g=1}^G \pi_g \mathcal{N}(t \mid \bar{t}_g,\Sigma_{tt,g})}
\end{equation}
is the nonlinear activation function.

\section{Experiments}\label{sec: experiments}
\begin{figure}[t]
    \centering
    \includegraphics[width=0.42\textwidth]{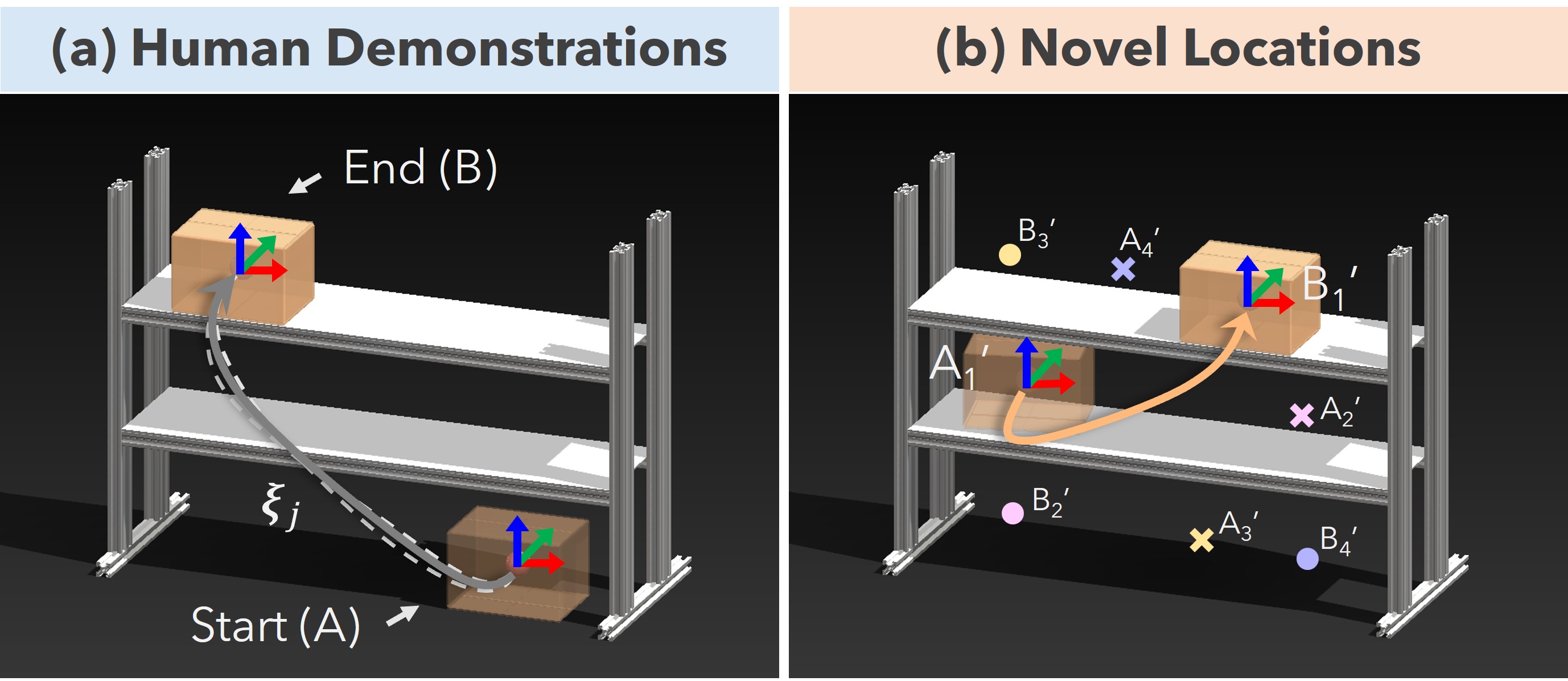}
    \caption{(a) Single-context demonstrations of grasp, transport, and release phases, where a box is transported from location A to location B. (b) The goal is to leverage these demonstrations to generate trajectories for any novel start and goal poses (A' and B'). Example pairs are shown, with one generated trajectory illustrated.}
    \label{fig: problem_setup}
\end{figure}

The proposed method is evaluated on a pick-and-place task in a shelf environment (Fig. \ref{fig: problem_setup}). The spatial constraints of the shelf limit feasible motions, making it an ideal scenario to test skill generalization while preserving key human manipulation strategies, such as careful extraction and reinsertion of objects. This setting emphasizes maintaining the fidelity of essential skill components for robust trajectory generation under novel configurations.

This experiment evaluates the method's ability to generalize from single-context demonstrations (Fig. \ref{fig: problem_setup}a) to diverse unseen task configurations (Fig. \ref{fig: problem_setup}b). It considers both extrapolation scenarios, which dominate due to single-context demonstrations, and interpolation cases, while ensuring boundary convergence and maintaining key characteristics of the demonstrated skills. This setup highlights the approach's effectiveness in learning and adapting from sparse, task-specific data.

We first outline the procedure for collecting human demonstrations and the parameter settings used for \g{GMM} training in Sec. \ref{sec: human_demonstration}. Next, Sec. \ref{sec: evaluation_protocol} details the evaluation protocol for comparing our method with baselines. Finally, Sec. \ref{sec: experiment_protocol} presents the real robot experiments executing the resulting generalized trajectories for the manipulation task.

\subsection{Human Demonstrations}\label{sec: human_demonstration}
\begin{figure}[t]
    \centering
    \includegraphics[width=0.48\textwidth]{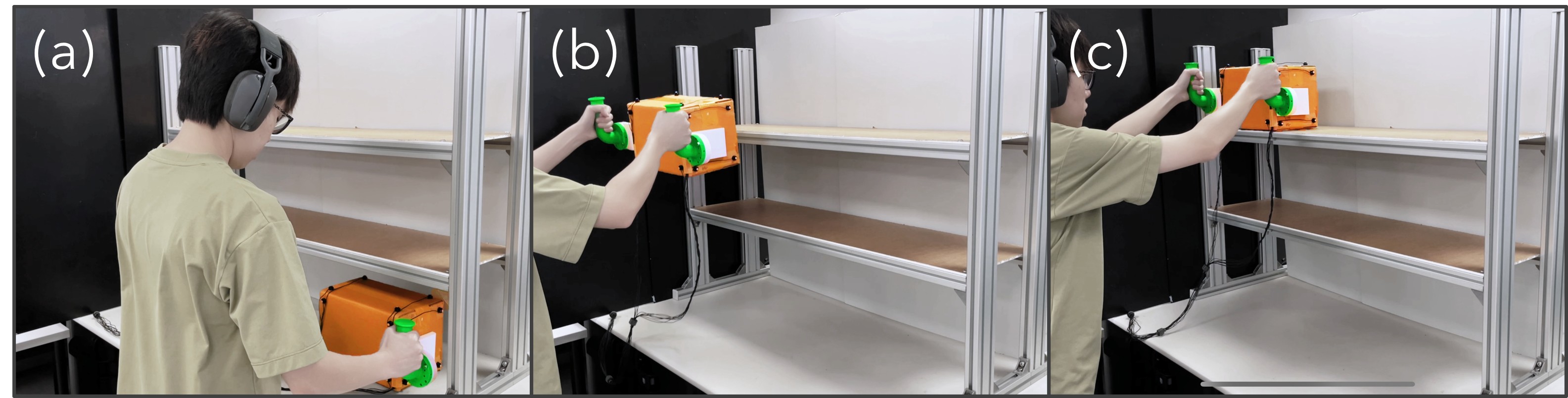}
    \caption{Sequential stages of human demonstration for the box pick-and-place task: (a) grasping, (b) transporting, and (c) releasing.}
    \label{fig: human_demonstration}
\end{figure}

We collect $\mathcal{J}=5$ demonstrations of the box pick-and-place task, all in a single-context with nearly identical start and goal poses. Each demonstration is represented as a trajectory $\V\xi_j$ (Sec. \ref{sec: gmm_training}), with 6-dimensional pose state $\V x_j \in \R^6$. Each trajectory comprises three sequential phases of fixed duration: (i) grasping the box (1 s), (ii) transporting it from location A to location B on the shelf (5 s), and (iii) releasing the box (1 s), as shown in Fig. \ref{fig: human_demonstration}. Task demonstrations are performed using two 3D-printed planar handles, one per hand. This approach enables teaching independent of the robotic system while mimicking the robot manipulation behaviors, offering flexibility and ease of demonstration while minimizing hardware dependencies and cost.

\begin{figure}[t]
    \centering
    \includegraphics[width=0.33\textwidth]{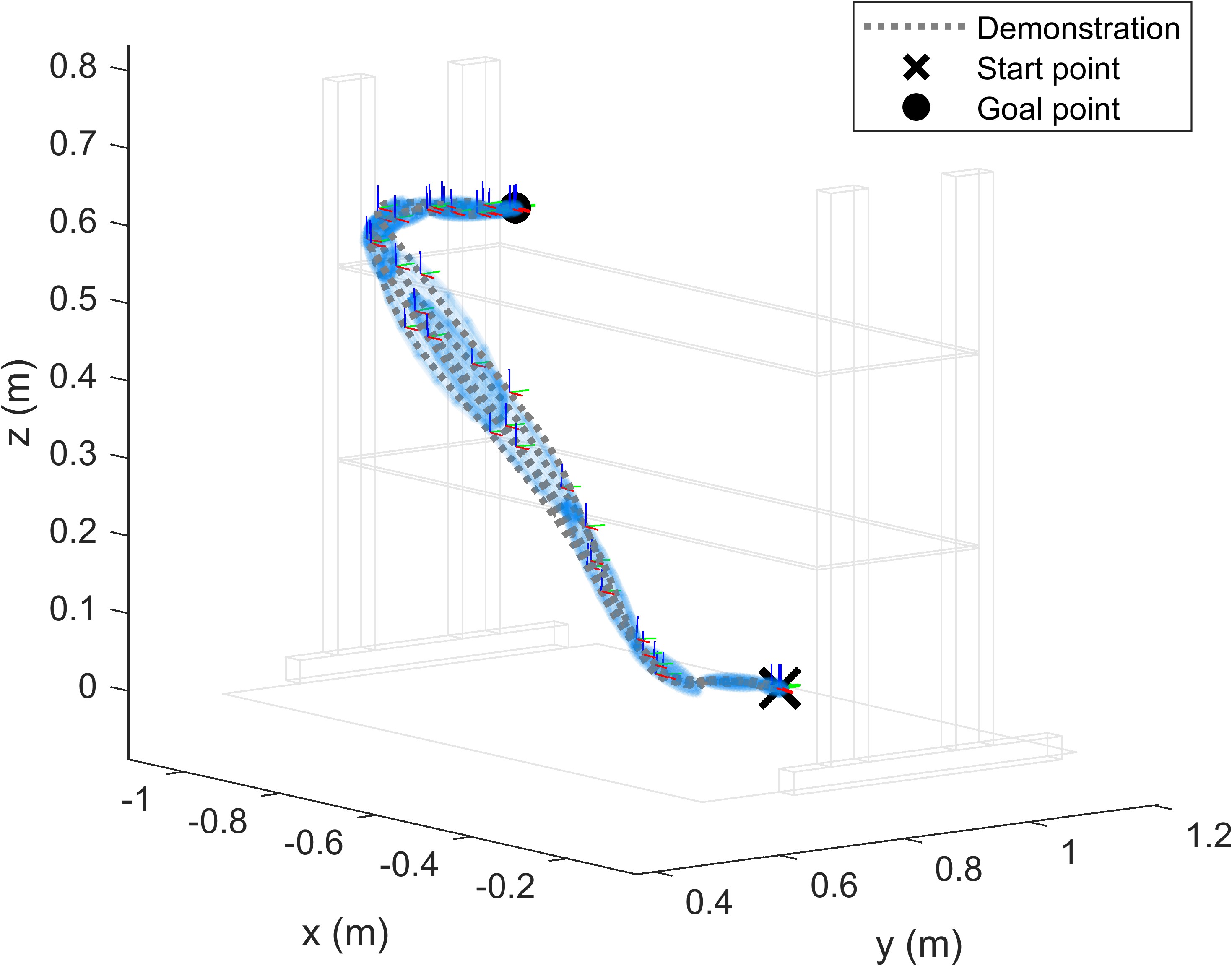}
    \caption{Demonstration trajectories (grey dotted) are shown alongside colored axes representing the orientation of the object during motion. The Gaussian components are visualized as blue ellipsoids.}
    \label{fig: demonstration_gmm}
\end{figure}

Given the symmetric coordination of the dual-arm motion, the manipulation behavior can be effectively represented by the box trajectory. Thus, box trajectories serve as a compact and sufficient representation of the demonstrated skills. Data were collected at 1000 Hz using a 3D motion capture tracker (VZ4000, Phoenix Technologies Inc.), with twelve active markers, in which four on each of the three box faces, tracking its pose along with timestamp. The demonstration data are modeled using a \g{GMM} with $G=15$ Gaussian components, yielding the resulting model shown in Fig. \ref{fig: demonstration_gmm}.

\subsection{Performance Evaluation}\label{sec: evaluation_protocol}
To assess generalization, we benchmark our method against widely used baseline algorithms, including \g{TP-GMM}, \g{DMP}, and \g{ProMP}, which are standard approaches for learning and reproducing robot manipulation skills from demonstrations.

We employ a progressive evaluation protocol that systematically increases task complexity under varying conditions:
\begin{enumerate}
    \item \textbf{Translational variations:} Each method is evaluated on novel start and goal poses where only the translational components vary, while the orientation components remains fixed. Specifically, the goal poses are parameterized along the shelf's length ($l$) and height ($h$):
    \begin{itemize}
        \item $l \sim \mathcal{U}(l_{min}, l_{max})$, covering the full available horizontal extent of the shelf.
        \item $h \in \{h_1,h_2,\dots,h_H\}$, covering the available vertical shelf levels. 
    \end{itemize}
    \item \textbf{Combined translational-rotational variations:} Building on the translational variations, this scenario introduces rotational variations around the vertical axis. Each method is evaluated on novel start and goal poses including rotations:
    \begin{itemize}
        \item $\theta \sim \mathcal{U}(-\pi/4, \pi/4)$, covering rotations within a quarter-circle range about the vertical axis.
    \end{itemize}
    Both translational and rotational variations are applied simultaneously to simulate more challenging task conditions.
\end{enumerate}
This staged evaluation first examines each method's spatial generalization, then its robustness when both position and orientation vary. For each variation, start and goal poses are randomly sampled from the distributions above, with all methods evaluated over 50 independent trials per variation type.

Performance is evaluated through trajectory-level generalization metrics:
\begin{itemize}
    \item \textbf{Trajectory success rate:} The proportion of generated trajectories that are feasible and collision-free, accounting for object dimensions.
    \item \textbf{Boundary pose error:} Computed independently for start and goal poses by comparing the target poses with the first and last poses of each generated trajectory. Translational error is the Euclidean distance between target and generated positions, while rotational error is the angular deviation between target and generated orientations. Errors are calculated across all generated trajectories, regardless of success.
    \item \textbf{Phase deviation:} Measures the average deviation of the generated trajectory from its mean pose during grasp and release phases. Translational deviation reflects position maintenance, and rotational deviation reflects orientation preservation. Lower values indicate closely adherence to the nearly constant demonstrated pose during these phases.
    \item \textbf{Shape deviation:} Measured using a Procrustes-based metric. Each generated trajectory is normalized for translation and scale, then optimally rotated and circularly shifted to align with the reference. The Procrustes distance quantifies deviation in global shape independent of location, orientation, scale, or start and goal points. Lower values indicate higher fidelity to the demonstrated trajectory, with values near zero corresponding to near-identical shapes.
\end{itemize}
Task-level measures are important in practice, but the present study focuses on trajectory-level evaluation. To complement this quantitative evaluation, we validate the feasibility of deploying the generalized trajectories on hardware through real-world dual-arm robotic experiments.

\subsection{Real-World Robotic Experiments}\label{sec: experiment_protocol}
In this section, a control framework for a dual-arm robotic system with two Kinova Gen3 7-DoF manipulators is presented, based on concepts from a recent framework \cite{campolo2025geometric}. Each end effector is equipped with a 3D-printed planar handle, identical to those used during human demonstrations.

\begin{table*}[t]
\centering
\caption{Performance comparison of different methods under training and randomized testing conditions. Training uses the average of demonstrations as target start-goal poses. Testing uses 50 randomized target start-goal poses, and results averaged. Metrics include trajectory success rate, boundary pose errors, phase deviation during grasp/release phases, and shape deviation (relative to the averaged demonstration trajectory in both cases). Values below 0.01 are shown as 0.00.}
\label{tab: result}
\begin{tabular}{l c cc cc c}
\toprule
\textbf{Method} & \textbf{Trajectory Success Rate (\%)} & \multicolumn{2}{ c }{\textbf{Boundary Pose Error (mm / deg)}} & \multicolumn{2}{ c }{\textbf{Phase Deviation (mm / deg)}} & \textbf{Shape Deviation} \\
\cmidrule(lr){3-4} \cmidrule(lr){5-6}
 &  & \textbf{Start} & \textbf{Goal} & \textbf{Grasp} & \textbf{Release} &  \\
 \midrule
\multicolumn{7}{l}{\textbf{Training Condition}} \\
\midrule
Ours & 100.0 & 0.89 / 0.05 & 0.33 / 0.27 & 5.51 / 0.32 & 1.17 / 0.07 & $1.65\times10^{-4}$ \\
\g{DMP} & 100.0 & 0.00 / 0.00 & 0.92 / 0.25 & 6.54 / 0.39 & 2.26 / 0.26 & $0.51\times10^{-4}$ \\
\g{TP-GMM} & 100.0 & 0.46 / 0.12 & 0.26 / 0.26 & 7.09 / 0.41 & 3.57 / 0.24 & $1.18\times10^{-4}$ \\
\g{ProMP} & 100.0 & 1.60 / 0.07 & 0.66 / 0.14 & 5.53 / 0.41 & 1.72 / 0.23 & $7.92\times10^{-4}$ \\
\midrule
\multicolumn{7}{l}{\textbf{Translational Variations}} \\
\midrule
\rowcolor{gray!10} \textbf{Ours} & \textbf{88.0} & 0.79 / 0.07 & 0.27 / 0.29 & 4.63 / 0.27 & 1.28 / 0.68 & 0.10 \\
\g{DMP} & 12.0 & 0.00 / 0.00 & 33.09 / 0.25 & 36.69 / 0.05 & 6.85 / 0.17 & 0.23 \\
\g{TP-GMM} & 20.0 & 425.73 / 0.19 & 425.83 / 0.29 & 6.74 / 0.35 & 4.17 / 0.83 & 0.00 \\
\g{ProMP} & 2.0 & 485.95 / 4.67 & 402.38 / 3.24 & 207.99 / 28.75 & 133.53 / 31.60 & 0.28 \\
\midrule
\multicolumn{7}{l}{\textbf{Combined Translational-Rotational Variations}} \\
\midrule
\rowcolor{gray!10} \textbf{Ours} & \textbf{78.0} & 0.79 / 0.07 & 0.27 / 0.29 & 4.61 / 0.50 & 1.29 / 1.53 & 0.11 \\
\g{DMP} & 8.0 & 0.00 / 0.00 & 34.38 / 1.17 & 38.08 / 1.22 & 7.11 / 0.29 & 0.24 \\
\g{TP-GMM} & 10.0 & 441.76 / 23.12 & 441.89 / 21.76 & 16.75 / 0.52 & 10.86 / 1.09 & 0.05 \\
\g{ProMP} & 0.0 & 527.80 / 16.10 & 417.93 / 11.05 & 482.03 / 72.28 & 465.44 / 91.62 & 0.50 \\
\bottomrule
\end{tabular}
\end{table*}

\begin{figure}[t]
    \centering
    \includegraphics[width=0.3\textwidth]{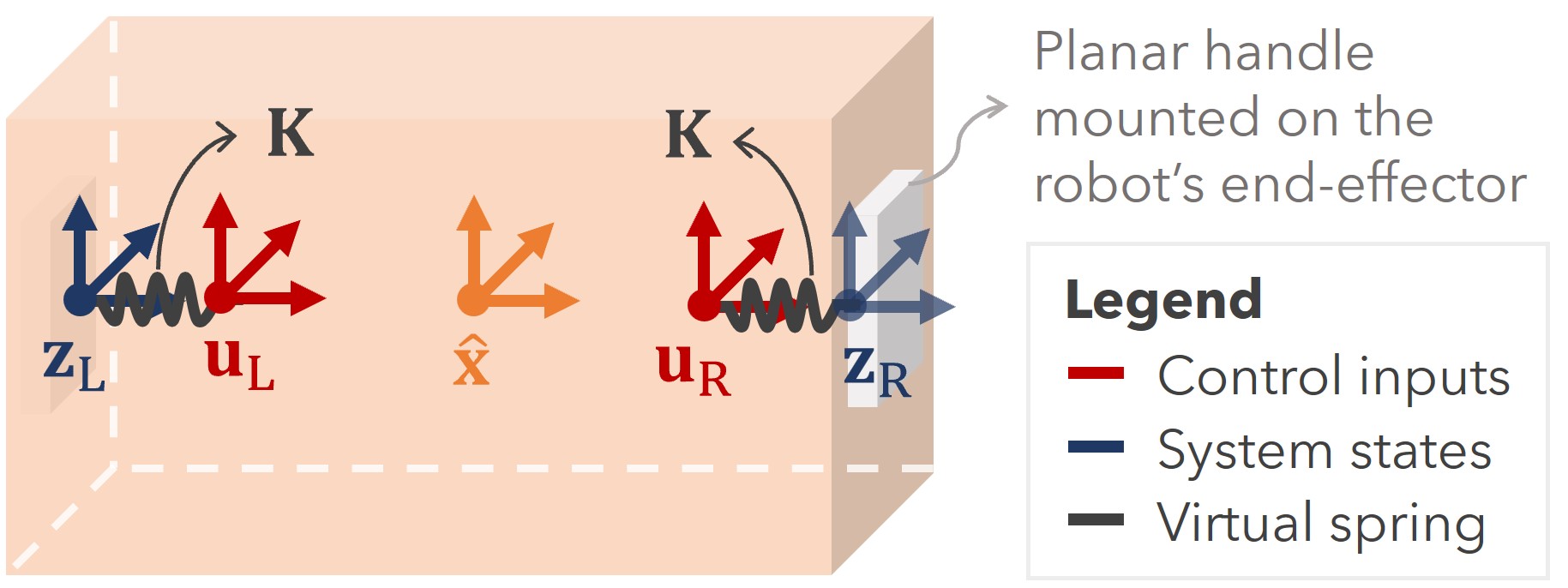}
    \caption{Control architecture of the dual-arm robotic system.}
    \label{fig: dual_arm_control}
\end{figure}

As shown in Fig. \ref{fig: dual_arm_control}, the dual-arm system is controlled via the control input $\V u = [\VT u_L, \VT u_R]^\T$, where $\V u_i \in \R^6$ denotes the control input for the robot $i$. The system state is defined as $\V z = [\VT z_L, \VT z_R]^\T$, where $\V z_i \in \R^6$ denotes the desired pose of the robot's handle $i$. These poses are computed from the generated box pose $\Vhat x \in \R^6$, with a positional offset equal to half the box width along the local box frame $\mathcal{B}$, while the handle orientation remains aligned with the box frame:
\begin{equation}
    \V z_i = \Vhat x + \bmat{s_i \frac{w}{2} \Vhat e_b \\ \V 0_3}, \quad s_L = -1, \ s_R = +1,
\end{equation}
where $\Vhat e_b \in \R^3$ is the unit vector along the box's local axis.\\
The relationship between $\V u_i$ and $\V z_i$ is governed by a nonlinear elastic behavior described by the stiffness matrix $\V K \in \R^{6\times 6}$. This can be conceptualized as a virtual spring connecting $\V u_i$ and $\V z_i$, expressed as:
\begin{equation}\label{eq: force}
    \V u_i = {\V K}^{-1}\V f_i + \V z_i
\end{equation}
where $\V f_i \in \R^6$ denotes the desired wrench applied at the robot handle $i$.\\
The applied wrench is modulated over three phases: grasp, transport, and release. During the grasp phase, the robot handles apply enough wrench to secure the box. In transport, the wrench is maintained to prevent slippage along the trajectory. Finally, during release, the wrench is gradually reduced to allow smooth disengagement of the handles.

To ensure that the robot follows the desired control input $\V u_i$, task-space impedance control is applied:
\begin{equation}\label{eq: forcebar}
    \Vbar f_i = \V K(\V u_i - \Vbar z_i)
\end{equation}
where $\Vbar z_i \in \R^6$ denotes the real-time handle pose feedback from robot $i$, and $\Vbar f_i \in \R^6$ is the resulting real-time wrench for robot $i$.\\
The task-space wrenches $\Vbar f_i$ are then converted into joint-space torques for the robot actuators using:
\begin{equation}\label{eq: torque}
    \Vbar \tau_i = \V J_{a,i}^\T \Vbar f_i
\end{equation}
where $\Vbar \tau_i \in \R^7$ is the commanded joint torque for robot $i$, and $\V J_{a,i} \in \R^{6\times 7}$ is the analytical Jacobian for robot $i$.\\
These torques $\Vbar \tau_i$ are sent directly to the robot's actuators, enabling execution of the desired manipulation while maintaining the desired task-space impedance and phase-dependent wrench modulation.

Following the execution of the control framework described above, only trajectories classified as successful according to the evaluation protocol (Sec. \ref{sec: evaluation_protocol}) were deployed on the physical platform. These experiments illustrate the practical feasibility of implementing the generalized trajectories on a physical system, rather than quantifying manipulation performance.

\section{Results and Discussion}\label{sec: experiment_result}
Table \ref{tab: result} summarizes the performance of our proposed method against three baselines (\g{DMP}, \g{TP-GMM}, \g{ProMP}) under both training and randomized testing conditions.

\textbf{Training condition.} For evaluation, the start and goal poses are set to the average of demonstrations, and shape similarity is measured with respect to the averaged demonstration trajectory. Under this setup, which closely reflects the training condition, all methods successfully reproduced the demonstration motions, achieving a 100\% trajectory success rate with minimal boundary pose errors, low phase deviation, and near-perfect shape similarity (on the order of $10^{-4}$), as shown in Table \ref{tab: result}. Since no qualitative differences were observed across methods, the training reproduction plots are omitted. These results validate our implementations and confirm that all methods can accurately reproduce demonstrations under training-like conditions.

\begin{figure*}[t]
    \centering
    \subfloat[] {
        \includegraphics[height=4.7cm]{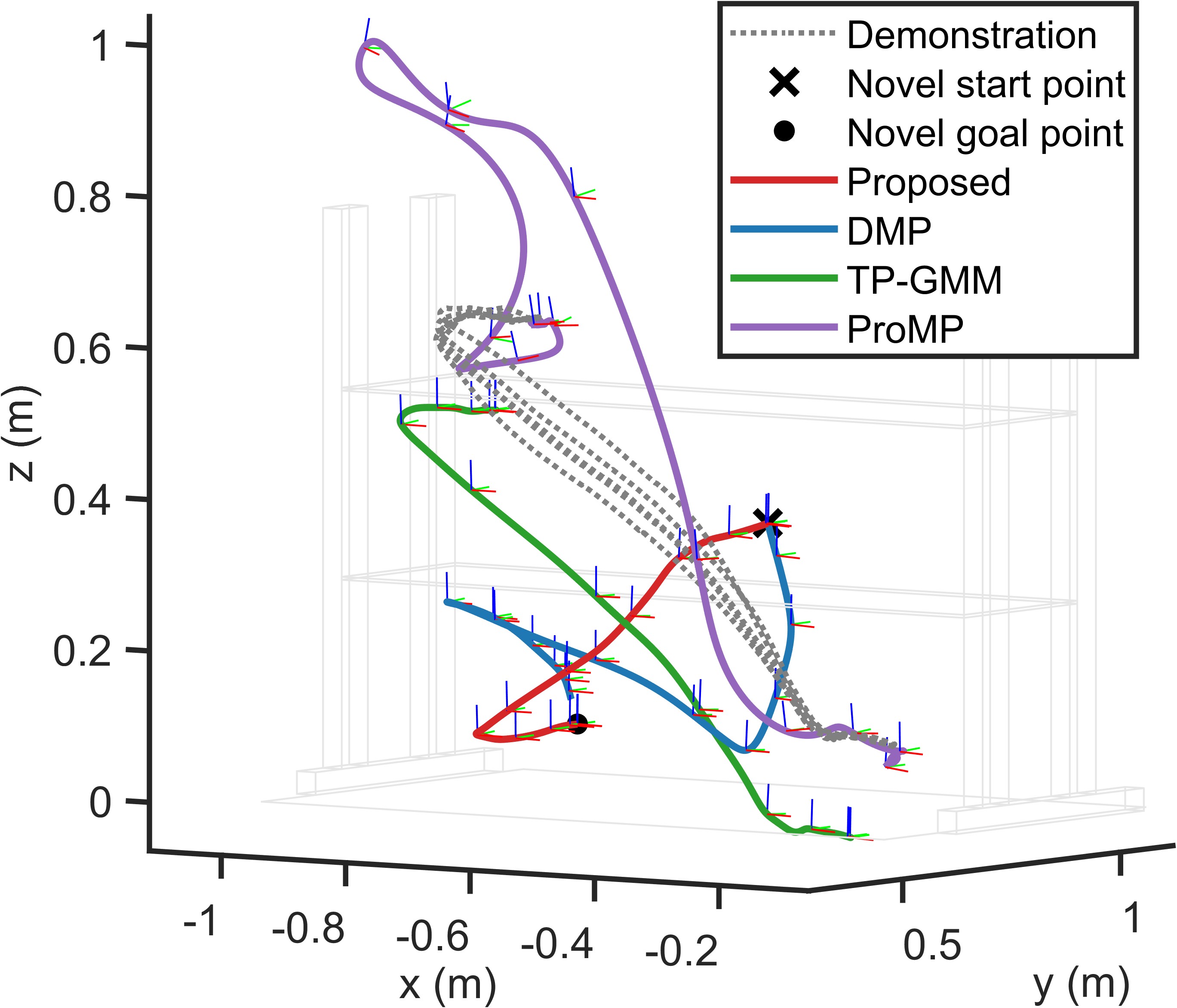}
        \label{fig: translation_3d}
    }
    \subfloat[] {
        \includegraphics[height=4.7cm]{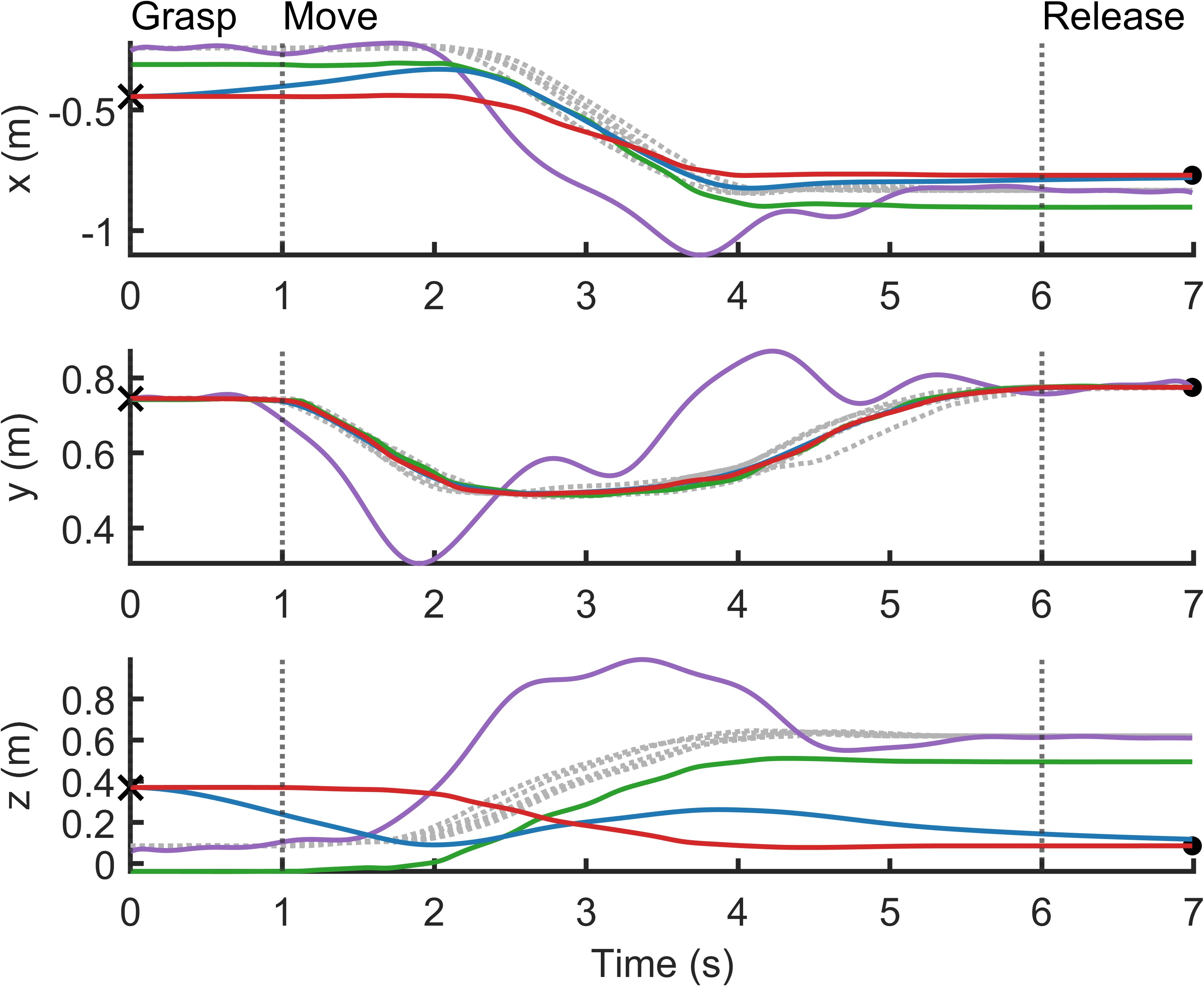}
        \label{fig: translation_2d_pos}
    }
    \subfloat[] {
        \includegraphics[height=4.7cm]{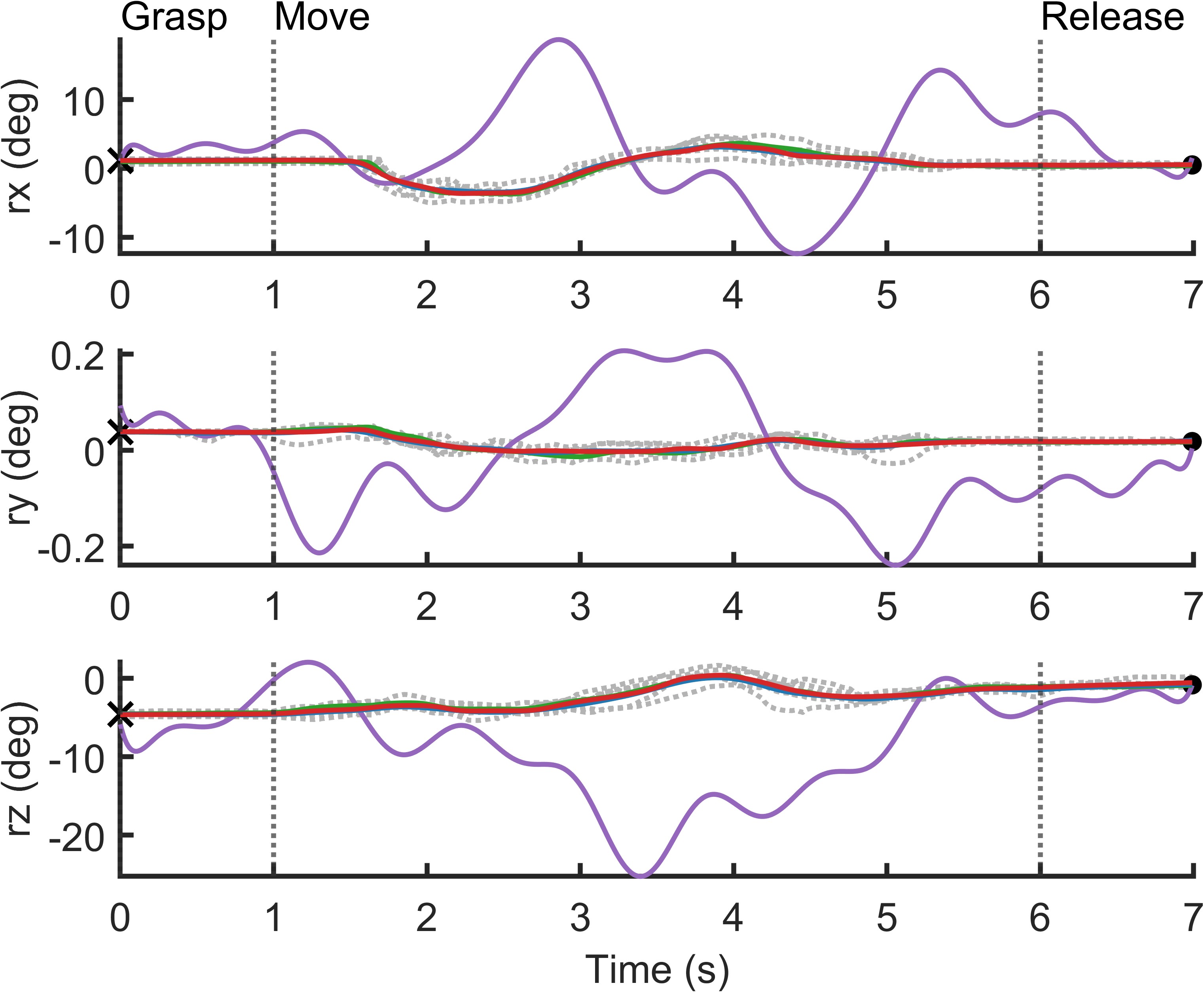}
        \label{fig: translation_2d_rot}
    }
    \caption{Representative trial under translational variations. (a) 3D trajectories for a novel start-goal configuration, with orientation frames shown at intervals. The shelf environment is shown for reference. (b) Position trajectories over time. (c) Orientation trajectories over time, represented as rotation vectors.}
    \label{fig: translation}
\end{figure*}

\textbf{Translational variations.} Figure \ref{fig: translation} shows representative trajectories generated by all methods under translational variations. From Table \ref{tab: result}, our method achieves an 88\% success rate, substantially outperforming all baselines ($\leq$20\%). As shown in Fig. \ref{fig: translation_3d}, our method generalizes successfully to unseen start-goal configurations (red trajectory), producing the desired motion patterns with consistent convergence at boundary points while maintaining phase-consistent motion. Failures occurred when generalizing bottom-to-top demonstrations to top-to-bottom tasks. The upward lift used to clear the shelf in the demonstrations produced premature downward motions when reversed, causing minor collisions. This suggests the method preserves spatiotemporal details, even when they do not align with inverted task contexts.
\begin{figure*}[t]
    \centering
    \subfloat[] {
        \includegraphics[height=4.7cm]{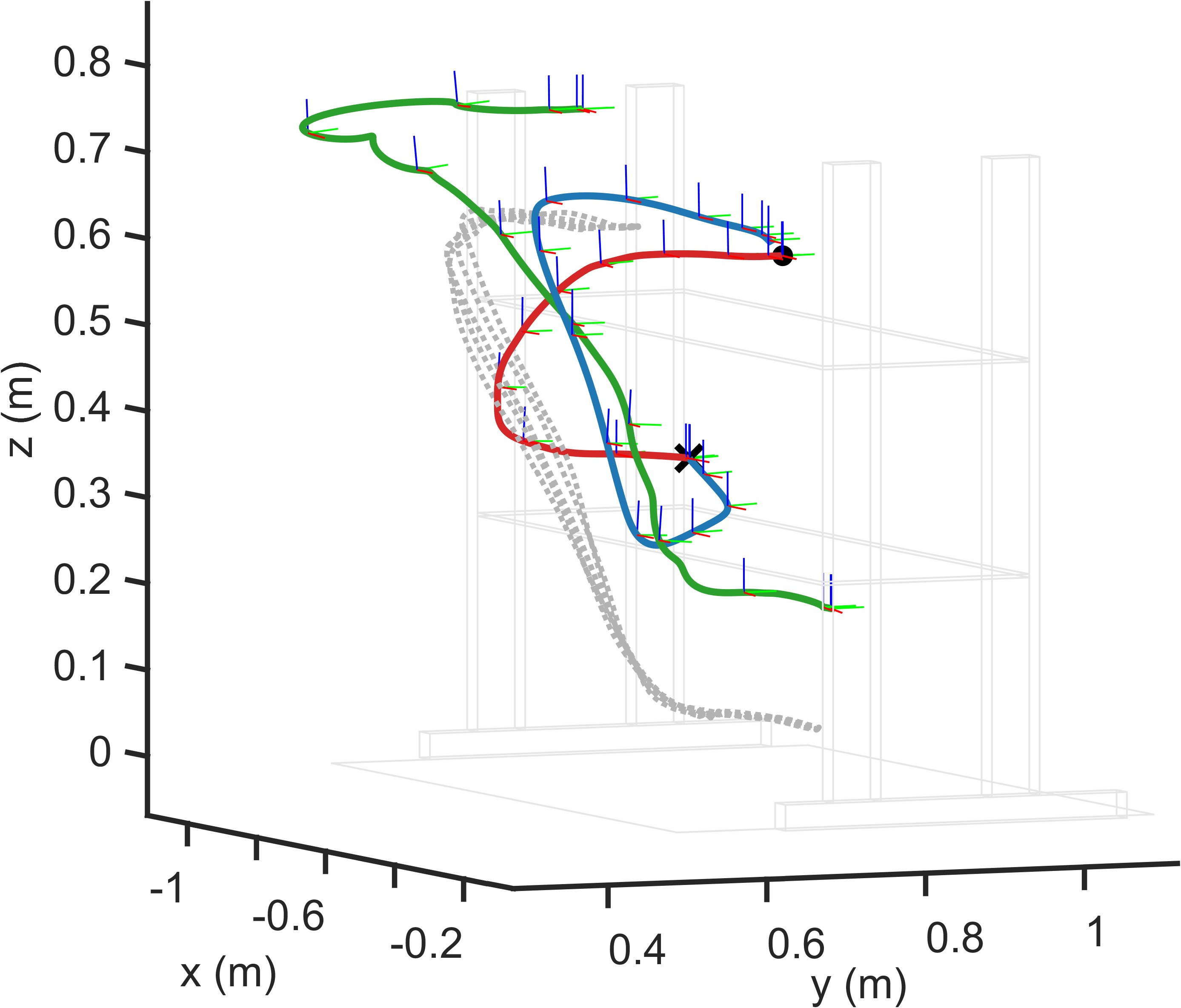}
        \label{fig: combined_3d}
    }
    \subfloat[] {
        \includegraphics[height=4.7cm]{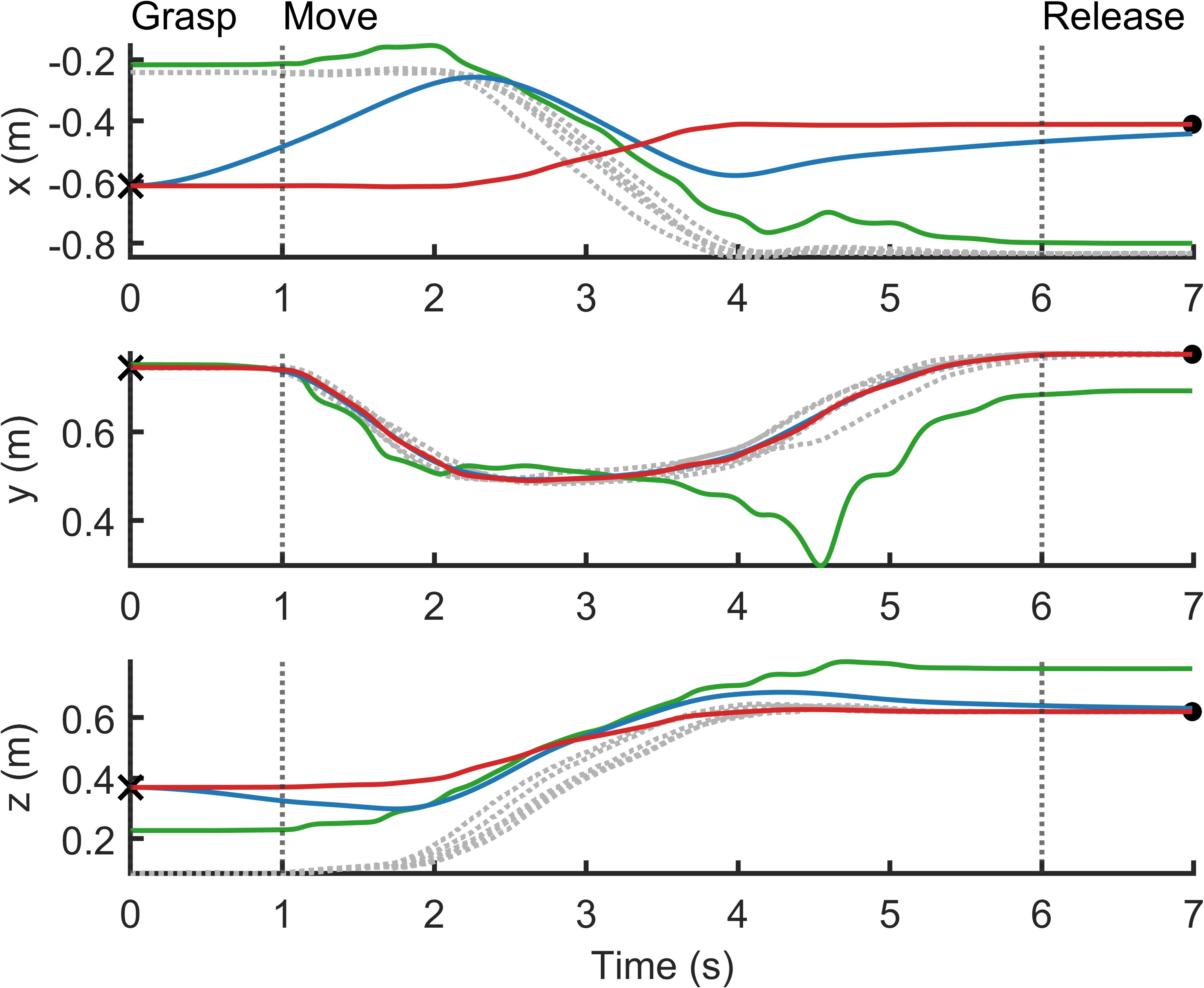}
        \label{fig: combined_2d_pos}
    }
    \subfloat[] {
        \includegraphics[height=4.7cm]{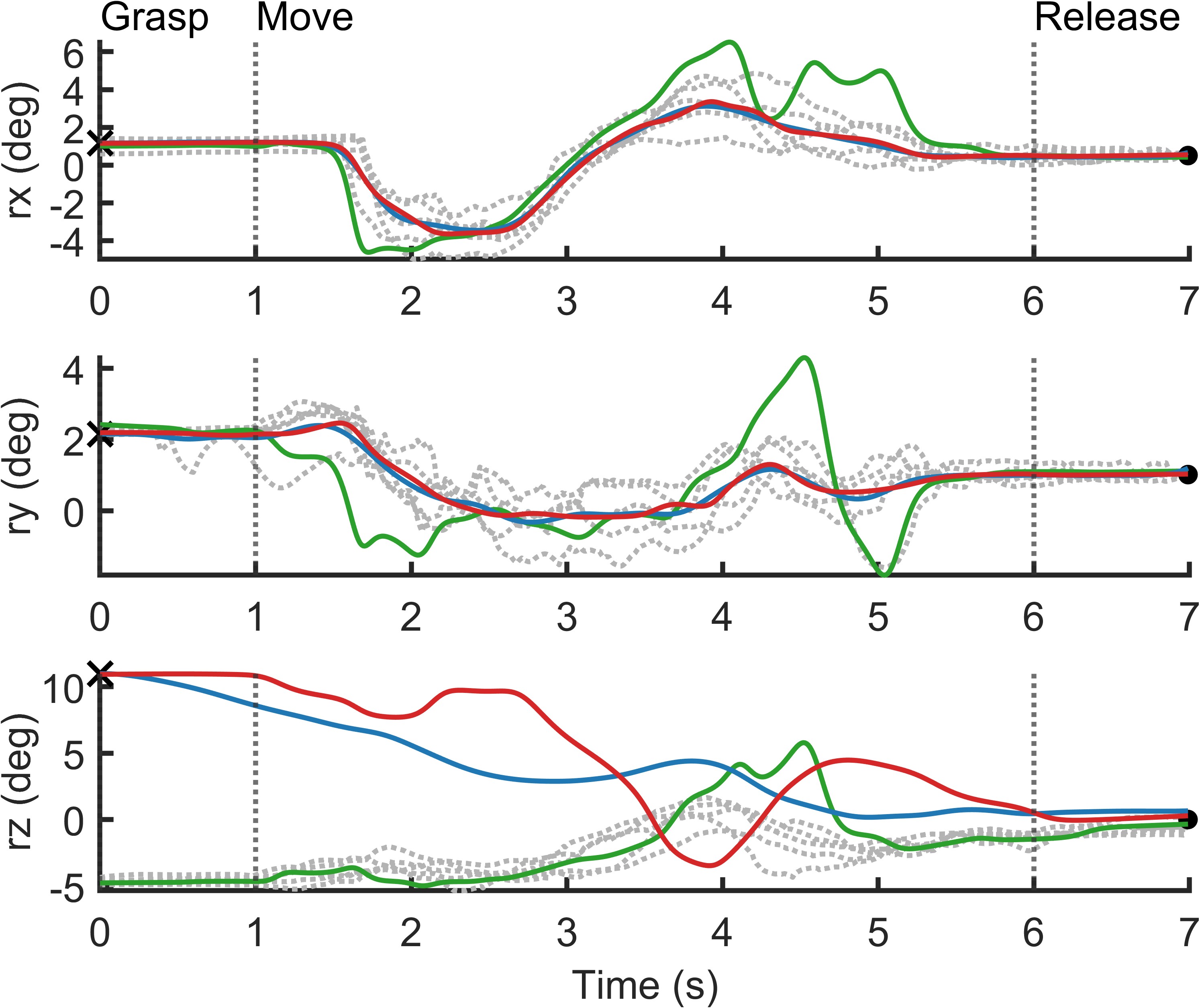}
        \label{fig: combined_2d_rot}
    }
    \caption{Representative trial under combined translational-rotational variations. (a) 3D trajectories for a novel start-goal configuration, with orientation frames shown at intervals. (b) Position trajectories over time. (c) Orientation trajectories over time as rotation vectors. \g{ProMP} results are omitted due to large deviations that obscure visualization.}
    \label{fig: combined}
\end{figure*}

In contrast, \g{DMP} exhibits only partial convergence: it reliably reaches the start point but diverges at the goal, with an average error of about 30 mm. More critically, the generated motion fails to reproduce the desired behavior, with a shape deviation score more than twice that of our method. These deviations reduce shape fidelity and ultimately cause task failures due to collisions with the environment (blue trajectory in Fig. \ref{fig: translation_3d}). \g{DMP} also does not explicitly capture phase-specific constraints, as reflected in its higher phase deviation values and drift along $x$ and $z$-axes during phases that should remain static (i.e., grasping and releasing) in Fig. \ref{fig: translation_2d_pos}. This limitations arises from \g{DMP} being a smooth goal-attractor system, where the nonlinear forcing term alone is insufficient to preserve fine-grained motion features when start-goal configurations deviate significantly from the demonstrations.

Both \g{TP-GMM} and \g{ProMP} fail to converge to the start or goal points. This arises from their probabilistic formulation: both require multi-context demonstrations to learn how trajectories adapt to varying conditions. With only single-context demonstrations, the learned distributions are too narrow, leading to poor generalization. \g{TP-GMM} tends to reproduce the demonstration shape (green trajectory in Fig. \ref{fig: translation_3d}), as seen in its low shape deviation and phase error, but it cannot adapt to the new start–goal configurations. As a result, trajectory success occurs only when test conditions closely resemble the demonstrations, while most cases fail due to lack of boundary convergence.

\g{ProMP}, in contrast, shows the most severe failures. Its generated trajectories neither converge to boundary points nor preserve the demonstrated motion structure (purple trajectory in Fig. \ref{fig: translation_3d}), instead producing incoherent trajectories. Figure~\ref{fig: translation_2d_rot} shows the orientation profiles: while all other methods remains close to the demonstrations, \g{ProMP} deviates substantially, confirming the loss of both shape fidelity and phase consistency. This combination of boundary points divergence and incoherent motion makes \g{ProMP} the weakest performer among the evaluated methods.

\textbf{Combined translational-rotational variations.} Introducing rotational variation alongside translational shifts further increases task difficulty. Our method remains robust under these conditions, achieving a 78\% success rate and significantly outperformed all baselines. As shown in Fig. \ref{fig: combined}, it continues to produce phase-consistent motions and converges to both start and goal configurations. Shape deviations remain low, indicating strong adherence to the demonstrated trajectory. While the angular phase deviation increase slightly, overall trajectory fidelity and task-relevant behaviors are largely preserved. In contrast, all baselines have their success rate dropped to $\leq$10\% and substantial increases in boundary pose errors, particularly in orientation.

\begin{figure}[t]
    \centering
    \includegraphics[width=0.48\textwidth]{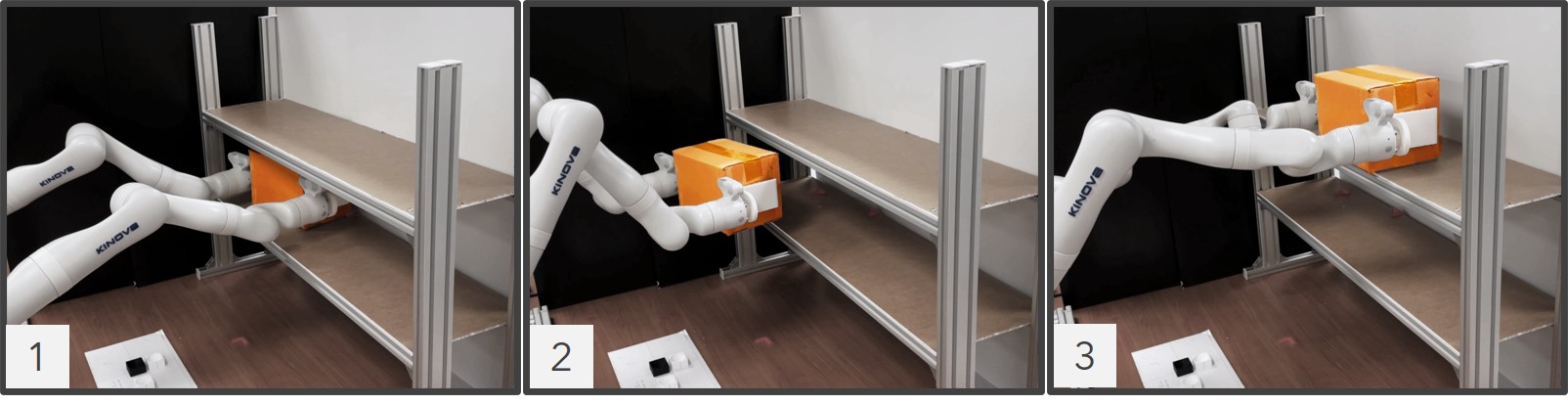}
    \caption{Key stages of real-world experiments with two Kinova Gen3 7-DoF manipulators: (1) grasping the box, (2) transporting it to the target location, and (3) releasing it.}
    \label{fig: real_world_results}
\end{figure}

Figure \ref{fig: real_world_results} shows key stages of the real-robot experiments, highlighting that the generalized trajectories generated by our method can be successfully executed.

\section{Ablation Study}
\begin{figure}[t]
    \centering
    \includegraphics[width=0.22\textwidth]{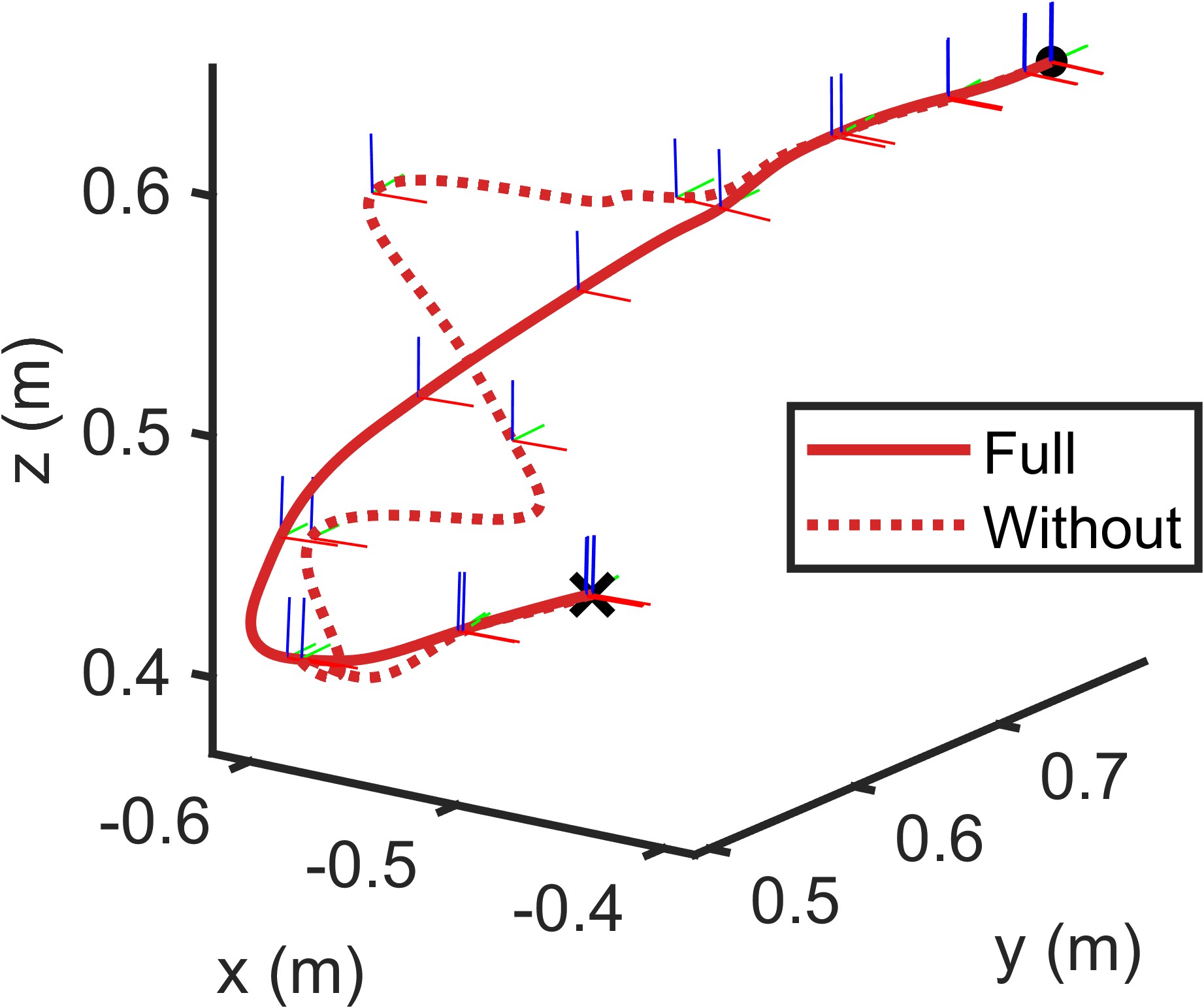}
    \caption{Comparison of generated trajectories with and without covariance reparameterization for a representative trial under combined variations, showing increased jerk when the module is removed.}
    \label{fig: ablation_study}
\end{figure}

An ablation study evaluate the impact of the covariance reparameterization module. As shown in Table \ref{tab: ablation}, disabling it reduces performance to 52\% and increases average jerk in both linear and angular. This highlights that covariance reparameterization is critical for generating smooth and physically feasible trajectories (Fig. \ref{fig: ablation_study}).
\begin{table}[t]
\centering
\caption{Ablation study on covariance reparameterization.}
\label{tab: ablation}
\begin{tabular}{l c cc}
\toprule
\textbf{Method} & \textbf{Success (\%)} & \multicolumn{2}{ c }{\textbf{Average Jerk}} \\
\cmidrule(lr){3-4}
&  & \textbf{Linear ($\mathbf{m/s^3}$)} & \textbf{Angular ($\mathbf{deg/s^3}$)}\\
\midrule
Full & 78.0 & 3.42 & 637.80 \\
Without & 52.0 & 10.16 & 802.94 \\
\bottomrule
\end{tabular}
\end{table}

\section{Conclusion and Future Work}\label{sec: conclusion}
In this work, we proposed a method for learning and generalizing demonstration-based trajectories from single-context demonstrations. By employing \g{GMM} as the learning framework, our approach enables rapid training from only a few human demonstrations, making it practical for capturing manipulation skills with minimal data. Beyond efficiency, the method robustly generalizes to unseen start-goal configurations while maintaining phase-consistent motion, accurate boundary points convergence, and high shape fidelity, outperforming baseline methods that often fail to reproduce the demonstrated motion structure or converge reliably under these conditions. The generalized trajectories can also be executed on a real robot, serving as qualitative validation of their practical feasibility.

While our method demonstrate strong generalization, some failures arise from minor collisions with the environment, indicating that incorporating collision-aware adaptation a promising direction for future work. One potential avenue is to model obstacles as repulsive potentials that locally modify the GMM parameters, enabling real-time trajectory adaptation. Beyond spatial generalization, a key direction for future work lies in temporal generalization. Our current approach assumes a fixed timing of motion, which can result in unnaturally fast or slow executions when the robot is required to traverse distances of varying scale. A natural extension would be to jointly adapt the temporal evolution of the trajectory alongside the spatial reparameterization. Such an approach would enable the robot to produce motions that are not only spatially accurate, but also temporally consistent and physically plausible across a broader range of task contexts.


\printbibliography

\end{document}